**Maxim Bazik**
*Vision Systems Inc.*
**Brien Flewelling**
*ExoAnalytic Solutions*
**Manoranjan Majji**
*Texas A&M*
**Joseph Mundy**
*Vision Systems Inc.*

## BAYESIAN INFERENCE OF SPACECRAFT POSE USING PARTICLE FILTERING

*Abstract:* Automated 3D pose estimation of satellites and other known space objects is a critical component of space situational awareness. Ground-based imagery offers a convenient data source for satellite characterization; however, analysis algorithms must contend with atmospheric distortion, variable lighting, and unknown reflectance properties. Traditional feature-based pose estimation approaches are unable to discover an accurate correlation between a known 3D model and imagery given this challenging image environment.

This paper presents an innovative method for automated 3D pose estimation of known space objects in the absence of satisfactory texture. The proposed approach fits the silhouette of a known satellite model to ground-based imagery via particle filtering. Each particle contains enough information (orientation, position, scale, model articulation) to generate an accurate object silhouette. The silhouette of individual particles is compared to an observed image. Comparison is done probabilistically by calculating the joint probability that pixels inside the silhouette belong to the foreground distribution and that pixels outside the silhouette belong to the background distribution. Both foreground and background distributions are computed by observing empty space. The population of particles are resampled at each new image observation, with the probability of a particle being resampled proportional to how the particle's silhouette matches the observation image. The resampling process maintains multiple pose estimates which is beneficial in preventing and escaping local minimums.

Experiments were conducted on both commercial imagery and on LEO satellite imagery. Imagery from the commercial experiments are shown in this paper.

### 1. INTRODUCTION

In the landscape of object pose research, the availability of enough textural information for feature extraction is a very common assumption. In this work, we consider the problem of pose estimation in the absence of satisfactory texture. We discover that a simple silhouette method when placed in a particle filter framework is sufficient to estimate pose for both conventional RGB imagery of aerial objects and ground-based telescope imagery of low earth orbit (LEO) objects with no reliance on texture.

Essential to this work are particle filters, a set of Monte Carlo algorithms for tracking the state of a dynamic system using Bayesian inference. Particle filtering has become an established method for solving many problems in vision and robotics including tracking, positioning, and navigation [2][3][6]. Particle filters are also used in many other areas such as economics [1], fault detection [5], and biological science [4]. The process of estimating the posterior distribution of the model parameters given the observed data is shown in Fig. 1.

Section 2 provides a more detailed discussion of particle filters. Section 3 discusses the technical details of the proposed approach. Section 4 shows the results of the proposed approach on two different data sources.

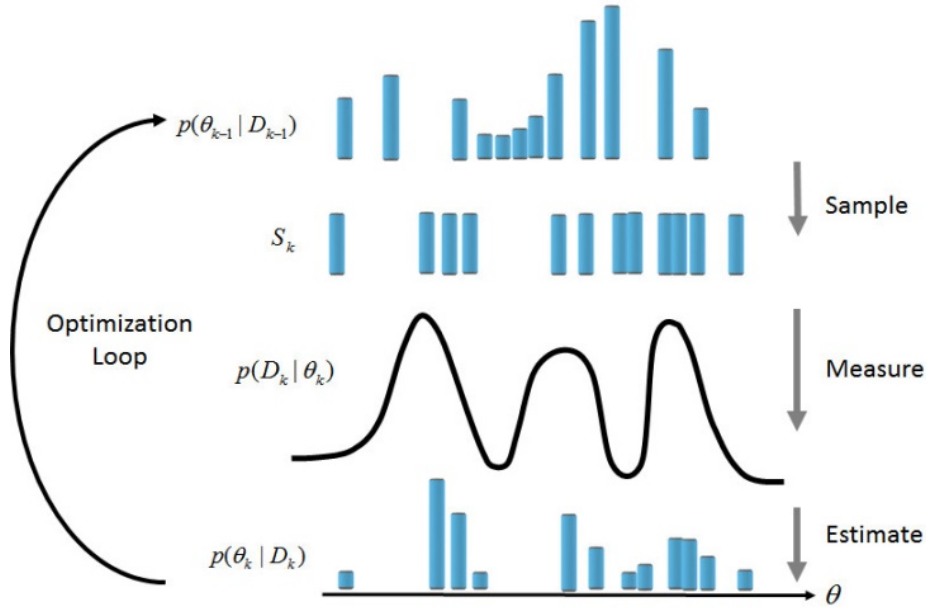

Fig. 1. The Particle Filtering Algorithm

## 2. OVERVIEW AND BACKGROUND

Put simply, particle filtering is a method to find the internal state of a system using inexact or incomplete measurements. The particle filter method works by hypothesizing many possible internal states, in particle filter nomenclature each hypothesized internal state is referred to as a particle. Each particle (or hypothetical internal state of the system) has a probability of being the actual state of the system.

Any particle filtering method requires some initialization; however, this initialization can be as trivial as sampling randomly from the set of all possible states. Sometimes the range of all possible states is far too large, and if some *a priori* information is known, the particles can be sampled from a constrained range of states. If still more *a priori* information is known, the initial particles can be sampled from a distribution of possible states based on some principled analysis of the probable state of the initial system. The method of initialization affects the final solution, for example an initialization close to the correct solution may converge quickly, while an initialization far from the correct solution may take much longer to converge or may be caught in local maxima.

Conveniently, the process for updating the particles to better reflect the actual system state is independent of the particle initialization method employed. This independence allows the discussion of the general process by which the particles are updated without further examination of the myriad initialization strategies one could employ; the specifics of the initialization method used in the experiments carried out are detailed in section 3 but can be summarize as random sampling from a bounded domain of possible states. So, upon initialization there exist many particles each of which represents one possible solution to the unknown internal state of the system. Coupled with each of these particles is the probability of being correct, that is the degree to which each particle reflects the system's actual state, at the current iteration, in probabilistic terms.

The algorithm moves beyond the initial set of particles and their associated probabilities when new information (in the form of an inexact or incomplete measurement) is provided. The process by which the algorithm moves from the previous state to the current state is usually termed the update process and is best described as a series of small steps. First particles are randomly sampled based on their probabilities, then each sampled particle is altered by a motion model to reflect the expected state at this current iteration. The motion model responsible for moving particles forward through time is probabilistic and problem specific. The motion model used for the experiments in this paper can be summarized as random jitters of the particle state, and there are various reasons for this the most obvious being the underdefined problem of determining the motion for a cameras system which is wobbling in an arbitrary manner. More detail on our motion model can be found in section 3. The probabilistic (and hence non-deterministic) nature of the motion model serves an auxiliary purpose, which is to facilitate the sampling of a particle multiple times in such a way that each sample produces a unique output particle. This production of these multiple unique outputs from a single input increases the space of possible solutions, when compared with a deterministic approach.

Once the particles are sampled and the motion of the sampled particles are projected into the current time step, the particles can be evaluated using the latest set of measurements. The evaluation of particles at this step is problem specific, typically done by using the known or approximated error distributions of the sensors to update the *a priori* probabilities of the particles. When the new particles and their corresponding probabilities are established the update-process is complete. The particles and their corresponding probabilities will not change again until new measurements are collected, these new measurements change the particles and probabilities by trigger the update process which always follows the same steps outlined in this section.

## 3. DETAILED TECHNICAL APROACH

The particle (or sample) $s_i$ is a parameter vector which encapsulate the hidden state of the system. The hidden state of the proposed pose estimation system is defined using six continuous variables: three variables encoding the yaw, pitch and roll of the object, two variables encoding the 2d translation of the object in the image domain, and one variable for object scale in the image domain. Object scale effectively encodes camera zoom, object distance, and object size in a single variable. These six degrees of freedom define a unique affine camera which can project a 3d object into a 2d space.

One objection to this approach would be to point out that the pose of the object is determined only relative to the pose of the camera. While true, this is only a problem when using imagery without metadata. For ground-based telescopes, the relative object pose can be transformed into absolute object pose using the well-known absolute camera pose.

As discussed in section 2, the particle filtering process requires some form of initialization. We assume limited *a priori* knowledge of the imagery and choose to initialize from a bounded distribution of random particles. The first bound is on the scale of the object, which cannot project larger than the size of the image or smaller than one tenth the size of the image. The second bound is on the translation of the objects, which is set so that at least half of the object must be visible in the image. These practical bounds were selected based on sample imagery and could be further tuned if necessary. The algorithm is bootstrapped by drawing a set of samples $S_0$ from this bounded random distribution of parameter vectors which is our prior distribution $p(\theta_0)$.

At the time step k, samples $S_k$ that are drawn from the current posterior distribution, $p(\theta_{k-1}|D)$, for the parameter vector given the observed data. A sample, or particle, is defined as a pair, $s_{ik} \in S_k = \{p_{ik}, \theta_{ik}\}$, i.e., a value for the parameter vector and an associated probability value. The motion model updates the parameter vector at each sample $m(\theta) = \theta'$. The initial distribution $p(\theta_0|D) \equiv p(\theta)$ is simply the prior distribution for the parameter vector. The samples are subjected to measurements from the observed data according to the distribution, $p(D|\theta'_k)$, which is the probability of the derivative image given the parameter set. Computing this distribution requires assigning each pixel of the target image to either the background $x \in b$ or foreground $x \in f$ as can be seen in the equation below. Pixels are assigned as the foreground satellite ($f$) or the empty star background ($b$) by rendering the satellite model using the pose information in the parameter vector $\theta'$ of sample $s_{ik}$. The background distribution $p(D_s(x)|x \in b)$ is empirical and is learned from images of the empty star background prior to observing the vehicle. The foreground or satellite distribution $p(D_s(x)|x \in f)$ can be simply taken as the uniform distribution. Thus, the probability of a measurement $D$ given the parameter vector $\theta_k$ is;

$$p(D|\theta'_k) = \frac{\prod_{x \in f} p(D_s(x)|x \in f) \prod_{x \in b} p(D_s(x)|x \in b)}{\prod_x p(D_s(x))} \quad (1)$$

The total data probability, $p(D_s(x))$, is simply the histogram of the entire image without segmentation into foreground and background. The denominator provides normalization with respect to the number of pixels in the overall image, so that the probability products are reasonably independent of resolution.

In practice the computation can be done very quickly for many particles by first computing the probability when all pixels are background;

$$\frac{\prod_x p(D_s(x)|x \in b)}{\prod_x p(D_s(x))} \quad (2)$$

Then by computing and caching the pixel-by-pixel foreground probability over the background probability;

$$c = \frac{p(D_s(x)|x \in f)}{p(D_s(x)|x \in b)} \quad (3)$$

Probabilities can be computed by multiplying over values in the cache. For efficiency, caching is done in the graphics processing unit (GPU) memory, and the final look up and mathematic operation are done on the GPU using the Open Computing Library (OpenCL) with the initial OpenGL resources passed directly to OpenCL without entering main memory. The equation computed is more accurately described by the equation below, but is equivalent to the one stated at the start of this section;

$$p(D|\theta_k') = \frac{\prod_x p(D_s(x)|x \in b)}{\prod_x p(D_s(x))} * \prod_{x \in f} \frac{p(D_s(x)|x \in f)}{p(D_s(x)|x \in b)} \quad (4)$$

The filtering process is very efficient also due to the GPU programming efforts. In Fig. 2, the probability calculation speeds are shown using the initial single threaded implementation, a multithreaded implementation and the current GPU based implementation. The efficiency of the evaluation pipeline could enable real-time performance if the parameter space is well bounded or an initial estimate close to ground truth is available.

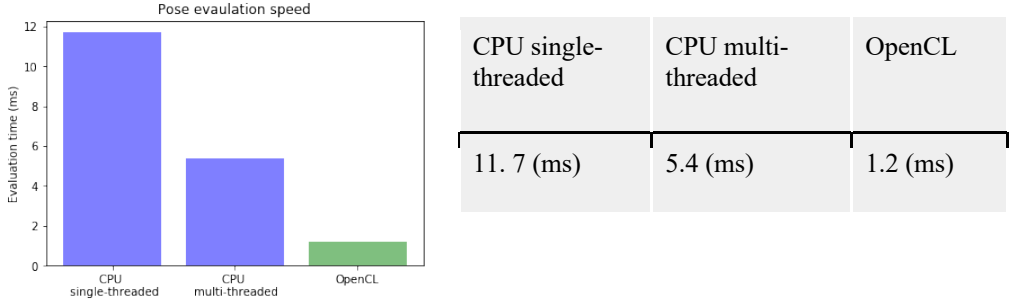

| CPU single-threaded | CPU multi-threaded | OpenCL |
|---|---|---|
| 11. 7 (ms) | 5.4 (ms) | 1.2 (ms) |

Fig. 2. The speed of per particle probability calculations

Once one update iteration has completed, the result is a new set of samples of the posterior distribution, $p(\theta_k|D)$, from which particles can be generated for the next iteration. The expected value of the parameter vector, $\bar{\theta}$, after any iteration, $k$, is given by,

$$\bar{\theta}_k = \frac{\sum_i p_{ik} \theta_{ik}}{\sum_i p_{ik}} \quad (5)$$

The procedure for sampling from the samples of $p(\theta_{k-1}|D)$ is carried out according to the following algorithm. A quantity, $c_k^i$, is defined, called the cumulative probability for sample $i$ at time step $k$. $c_k^i$ is defined by:

$$\begin{cases} c_k^0 = 0, \\ c_k^i = c_k^{i-1} + \frac{p_{ik}}{\sum_i p_{ik}} \end{cases} \quad (6)$$

That is, $c_k^i$ is computed by just accumulating the sample probabilities, a discrete form of the cumulative distribution. The sample set, $S_k$ is formed from the prior sample set $S_{k-1}$ as follows:

1. For each new sample, $i$, generate a random number $r$ on the interval, $[0 \quad 1]$ according to a uniform distribution
2. Find the smallest index, $j$, such that $c_{k-1}^j \geq r$
3. The sample, $s_{i,k} = s_{j,k-1}$

Particle filtering can be computationally expensive since the number of particles grows exponentially with the dimensionality of the parameter vector. On the other hand, since the posterior distribution is likely to be highly multi-modal, random particle generation avoids getting stuck at inferior local probability maxima.

## 4. EXPERIMENTS AND RESULTS

Model based pose estimation was conducted in many experiments. Early experiments used RGB imagery of a NASA shuffle returning to earth. Later experiments were conducted using imagery of Seasat in low earth orbit (LEO). In the latter case articulation of the main solar panel were added demonstrating the ease with which a new variable can be added to the space of samples.

The RGB results are found using off the shelf YouTube video of a shuttle reentering the atmosphere. The model used in pose estimation shown in Fig. 4 is an official NASA model.

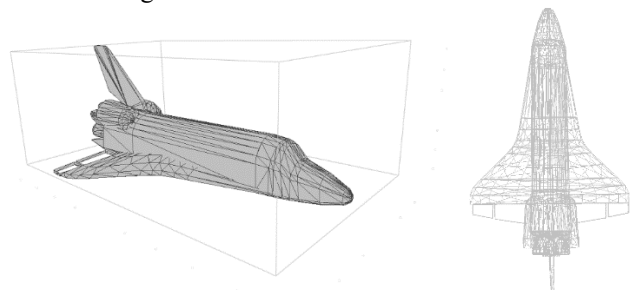

Fig. 4. The NASA Shuttle Model Used in the Pose Estimation Pipeline

The first row of Table 1 shows the input imagery, while the second shows the highest probability particle. In the final row the particle with highest probability is shown in blue overlaying the original image. The initial particles are distributed uniformly in the parameter space, but quickly converge to the correct solution

Table 1. The 1$^{st}$, 5$^{th}$, 10$^{th}$, and 100$^{th}$ Frame of the RGB Shuttle Sequence

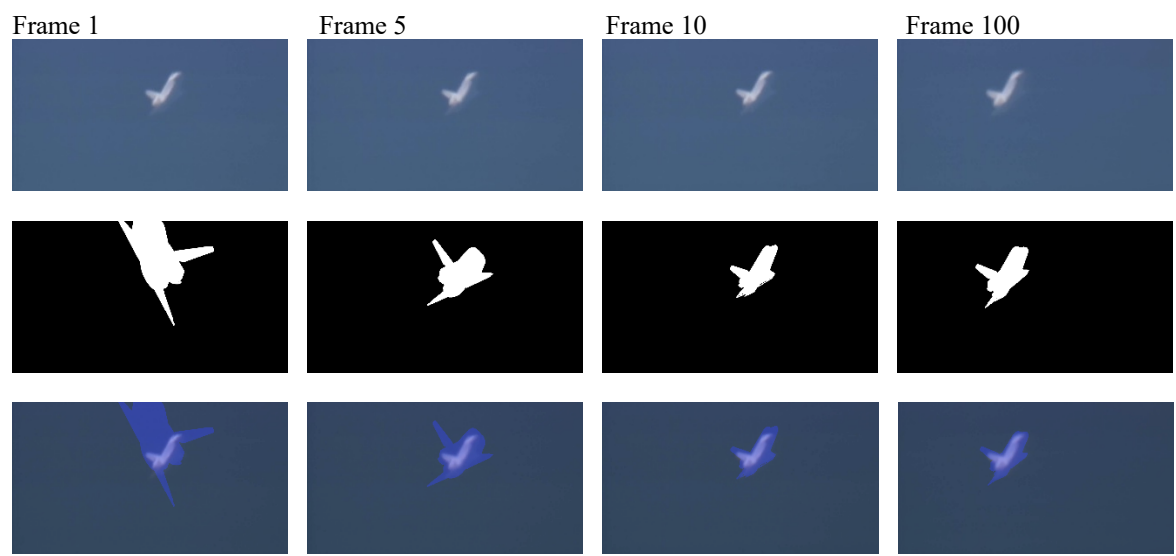

| Frame 1 | Frame 5 | Frame 10 | Frame 100 |

The Seasat experiments were conducted using imagery provided by the AFRL. These experiments showed comparable accuracy for Seasat in LEO, however these images are distribution limited and are not shown.

## 5. DISCUSSION

In this work we developed a texture independent pose estimation pipeline which relies on Bayesian inference to estimate pose. We show the validity of this approach by testing on two very different image sequences. The results shown in section 4 offer visual proof that the pose estimates software results in high quality pose estimates given enough frames. Secondly, we test the applicability of this approach in two very different image domains providing evidence for the generalizability of this method.

# 6. ABREVIATION AND ACRONYMS

1. AFRL        Air Force Research Lab
2. CPU         Central Processing Unit
3. GPU         Graphics Processing Unit
4. NASA        Nation Aeronautics and Space Administration
5. OpenCL      Open Computing Language
6. OpenGL      Open Graphics Library
7. VSI         Vision Systems Inc.